\documentclass[letterpaper]{article} 
\usepackage{aaai25}  
\usepackage{times}  
\usepackage{helvet}  
\usepackage{courier}  
\usepackage[hyphens]{url}  
\usepackage{graphicx} 
\usepackage{natbib}  
\usepackage{caption} 
\usepackage{algorithm}
\usepackage{algorithmic}

\usepackage{newfloat}
\usepackage{listings}
\DeclareCaptionStyle{ruled}{labelfont=normalfont,labelsep=colon,strut=off} 
\floatstyle{ruled}
\newfloat{listing}{tb}{lst}{}
\floatname{listing}{Listing}
\nocopyright 

\usepackage{booktabs}
\usepackage{multirow}

\usepackage{color}

\usepackage{subcaption}

\title{MM-CamObj: A Comprehensive Multimodal Dataset \\for Camouflaged Object Scenarios}
\author{
    Jiacheng Ruan\textsuperscript{\rm 1}\equalcontrib,
    Wenzhen Yuan\textsuperscript{\rm 1}\equalcontrib, 
    Zehao Lin\textsuperscript{\rm 2}, 
    Ning Liao\textsuperscript{\rm 2}, 
    \\
    Zhiyu Li\textsuperscript{\rm 2\dag}, 
    Feiyu Xiong\textsuperscript{\rm 2}, 
    Ting Liu\textsuperscript{\rm 1}, 
    Yuzhuo Fu\textsuperscript{\rm 1}\thanks{Zhiyu Li and Yuzhuo Fu are co-corresponding authors.}
}
\affiliations{
    \textsuperscript{\rm 1} Shanghai Jiao Tong University 
    \\
    \textsuperscript{\rm 2} Institute for Advanced Algorithms Research, Shanghai. 
    \\
    jackchenruan@sjtu.edu.cn

}

\usepackage{bibentry}

\begin{document}

\maketitle

\begin{abstract}


Large visual-language models (LVLMs) have achieved great success in multiple applications. However, they still encounter challenges in complex scenes, especially those involving camouflaged objects. This is primarily due to the lack of samples related to camouflaged scenes in the training dataset. To mitigate this issue, we construct the MM-CamObj dataset for the first time, comprising two subsets: CamObj-Align and CamObj-Instruct. Specifically, CamObj-Align contains 11,363 image-text pairs, and it is designed for VL alignment and injecting rich knowledge of camouflaged scenes into LVLMs. CamObj-Instruct is collected for fine-tuning the LVLMs with improved instruction-following capabilities, and it includes 11,363 images and 68,849 conversations with diverse instructions. Based on the MM-CamObj dataset, we propose the CamObj-Llava, an LVLM specifically designed for addressing tasks in camouflaged scenes. To facilitate our model's effective acquisition of knowledge about camouflaged objects and scenes, we introduce a curriculum learning strategy with six distinct modes. Additionally, we construct the CamObj-Bench to evaluate the existing LVLMs' capabilities of understanding, recognition, localization and count in camouflage scenes. This benchmark includes 600 images and 7 tasks, with a total of 9,449 questions. Extensive experiments are conducted on the CamObj-Bench with CamObj-Llava, 8 existing open-source and 3 closed-source LVLMs. Surprisingly, the results indicate that our model achieves a 25.84\% improvement in 4 out of 7 tasks compared to GPT-4o. Code and datasets will be available at https://github.com/JCruan519/MM-CamObj.



\end{abstract}
\section{Introduction}

Benefiting from the revolutionary breakthroughs of large language models (LLMs) \cite{minaee2024large,llama2,llamamoe,memory3,lvlmsothers1} and the continuous advancements in visual foundational models \cite{li2024multimodal}, large vision-language models (LVLMs) have achieved remarkable progress \cite{li2024multimodal}. Via training processes such as vision-language alignment \cite{zhu2023minigpt} and instruction fine-tuning \cite{lin2024vila}, LVLMs have demonstrated outstanding performance in various tasks, including image description, visual question answering, object localization, and visual reasoning. These achievements not only showcase the flexibility and adaptability of LVLMs in multimodal scenarios but also drive their widespread applications in fields such as medical imaging \cite{li2023artificial}, remote sensing \cite{li2024vision}, and autonomous driving \cite{cui2024survey}.

\begin{figure}
    \centering
    \includegraphics[width=0.95\linewidth]{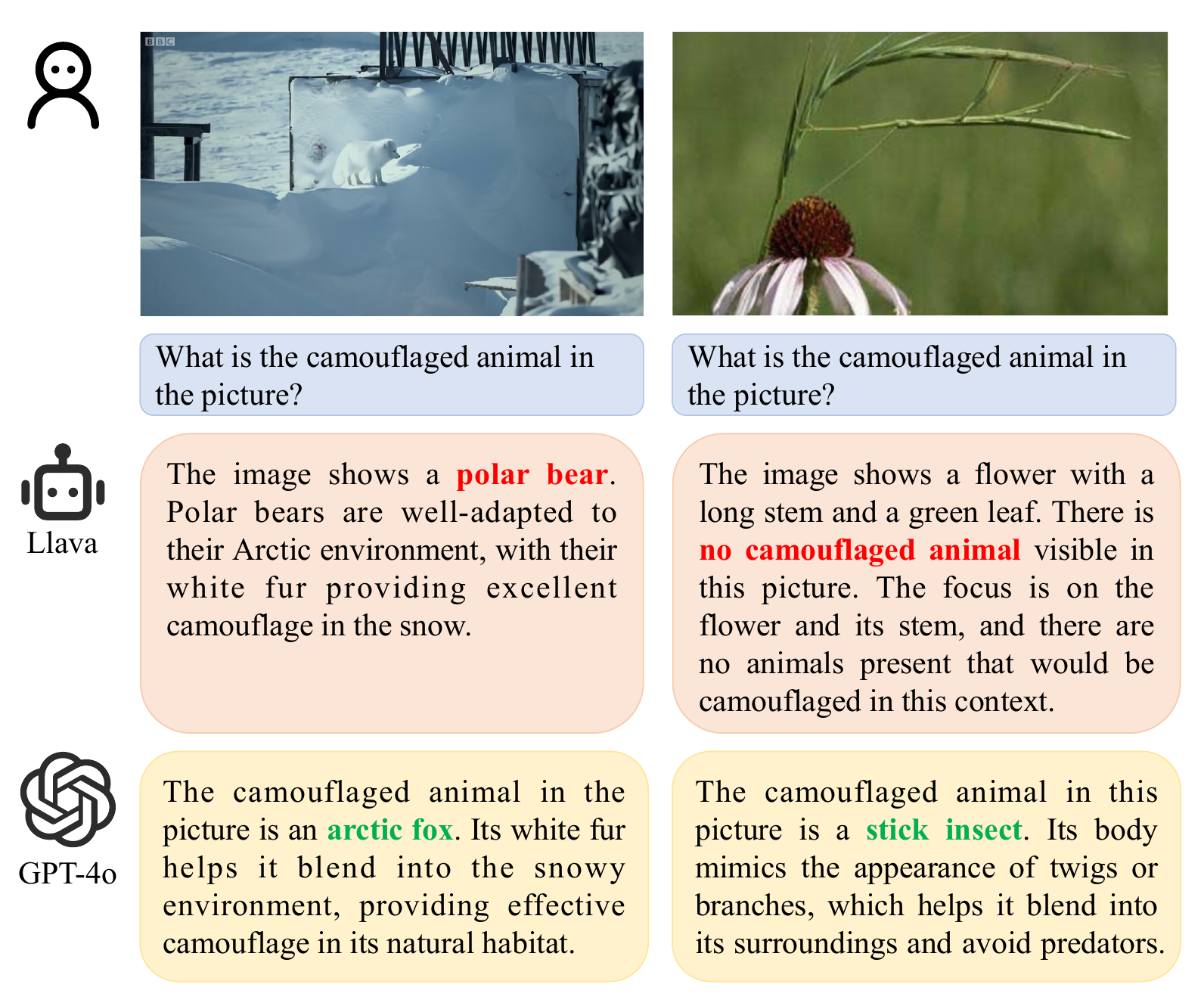}
    \caption{Empirical studies in camouflaged scenarios. We compare Llava-v1.5-7b and GPT-4o in understanding the two classical images with camouflage object existed. There is a certain gap between existing open-source models and advanced closed-source models in camouflage scenes.}
    \label{fig:head}
\end{figure}

Despite the significant progress made by LVLMs in various multimodal applications, they still suffer from significant challenges in complex scenes, particularly those involving camouflaged objects. The primary cause of the difficulty and complexity of tasks in such scenarios lies in the high similarity between the camouflaged objects and their surroundings in terms of texture, shape, and color. As illustrated in Figure \ref{fig:head}, our empirical experiments provide a preliminary demonstration of the performance of LVLMs in these scenarios. On the left side of Figure \ref{fig:head}, Llava misidentifies an Arctic fox as a polar bear, whereas on the right side, Llava fails to recognize a camouflaged stick insect. In contrast, GPT-4o accurately identifies both camouflaged objects. This indicates that existing open-source LVLMs still have shortcomings in understanding scenes with camouflaged objects and lag behind closed-source models. Similar observations can be found in ~\cite{eval-cod-lvlm, mmt-bench}.

This gap primarily stems from the lack of training samples related to camouflage scenes. To address this issue, we meticulously collect 11,393 images from four classical datasets for camouflage scene understanding \cite{ovcamo,COD10K,MoCA-Mask,PlantCamo,CPD1K}, and leverage the advanced GPT-4o to construct a semantic-rich and feature-diverse dataset called MM-CamObj. This dataset comprises two subsets: CamObj-Align and CamObj-Instruct, which are utilized in the alignment and instruction fine-tuning training phases, respectively. Specifically, CamObj-Align contains 11,393 high-quality image-text pairs, with texts providing detailed descriptions of the camouflage scenes in the images. This subset equips LVLMs with extensive knowledge of camouflage objects and scenarios. CamObj-Instruct consists of 11,393 images with 68,849 diverse conversations, where each image is assigned with at least three high-quality instructions and corresponding responses, and it can further enhance the instruction-following capabilities of LVLMs in camouflage scenarios.

To achieve superior performance in camouflage scene tasks, we propose a novel LVLM, dubbed CamObj-Llava, by further training Llava-v1.5 \cite{llava1_5} on our MM-CamObj. However, due to the difficulty of obtaining camouflage images, the training samples in MM-CamObj are relatively few compared to the large-scale datasets used in general scenarios, such as ALLaVA-Instruct-4V \cite{allava} with 715K samples and Bunny-695K \cite{bunny}. This condition of limited samples poses significant challenges to the model's training. To address this issue, we introduce a curriculum learning strategy \cite{curriculum_learning_survey_2}. It enhances the model's capabilities by adopting a learning sequence from easy to difficult, allowing the model to first learn simple samples and then gradually encounter complex ones. In this study, we rank the samples in MM-CamObj according to six different evaluation modes and input them into the LVLM for training in order of increasing difficulty. This strategy enables CamoObj-Llava to acquire a more comprehensive understanding of camouflage scenes, improving its performance in complex camouflage scenarios.

To comprehensively evaluate the performance of LVLMs in camouflage scenarios, we introduce the CamObj-Bench, a benchmark specifically designed to assess LVLMs' performance in multimodal camouflage tasks. CamObj-Bench comprises 600 meticulously curated images, covering 144 species across diverse biological categories, including aquatic animals, mammals, birds, insects, plants, etc., greatly enhancing the richness and diversity of the benchmark. We have established seven core tasks—Easy VQA, Hard VQA, Bbox Location, Image Caption, Count Choice, Mask Match, and Mask TF—to thoroughly evaluate the capabilities of LVLMs in camouflage scenarios. Each task is rigorously designed to ensure diversity and challenge, posing difficulties even for powerful models like GPT-4o. CamObj-Bench serves not only as a comprehensive tool for evaluating LVLM performance in multimodal camouflage tasks but also provides valuable insights for future advancements in this field.

The main contributions are summarized as follows:

\begin{itemize}
    \item \textbf{MM-CamObj.} We have developed the MM-CamObj dataset to enhance the existing LVLM's capabilities in camouflaged object scenarios. This dataset includes two subsets: CamObj-Align and CamObj-Instruct, utilized for the alignment and instruction-tuning stages of LVLM training, respectively.
    \item \textbf{CamObj-Llava.} We have developed CamObj-Llava, an LVLM specifically designed for tasks in camouflaged scenes. By leveraging diverse samples from MM-CamObj dataset and six different curriculum learning methods, CamObj-Llava exhibits outstanding performance across various downstream tasks.
    \item \textbf{CamObj-Bench.} We have established CamObj-Bench, a benchmark aimed at comprehensively and systematically evaluating the performance of LVLMs in camouflaged object scenarios. This benchmark consists of 7 tasks, with a total of 9,449 carefully designed questions.
\end{itemize}

\section{Related Works}
\subsection{Camouflaged Scene Understanding (CSU) Datasets}






In nature, animals use various ingenious camouflage techniques to hide themselves and avoid being detected by predators. Since these camouflage methods are often highly complex, accurately understanding camouflaged scenes and identifying camouflaged objects is a challenging task. Many diverse datasets~\cite{CAD,CPD1K,COD10K,MoCA-Mask,ovcamo,PlantCamo} has propelled the development of the CSU field. For example, COD10K~\cite{COD10K} is a large-scale dataset for camouflaged object detection that covers 78 categories, aiming at identifying objects that visually blend into the background. PlantCamo~\cite{PlantCamo} is an image dataset focused on camouflaged plants, containing images of various camouflaged plants. Each image is accompanied by its corresponding species label and includes mask annotations to display the plant's contour. CPD1k~\cite{CPD1K} is a dataset specifically constructed for the task of camouflaged personnel detection. The dataset consists of images extracted from multiple video clips captured in different natural scenes, covering a variety of camouflage patterns and complex lighting and occlusion conditions. MoCA-Mask~\cite{MoCA-Mask} is a large-scale dataset for video camouflaged object detection. This dataset provides pixel-level manually annotated masks for every five frames of the video to show the contour of camouflaged objects.


\subsection{Large Vision and Language Models}

Large vision language models (LVLMs) are built upon large language models (LLMs)~\cite{llama,llama2,llama3,vicuna,qwen,qwen2} and visual encoders~\cite{openaiclip,evaclip,dinov2,sigmoidclip}. Through visual alignment and visual instruction tuning, they effectively fuse image and text information, enhancing performance in multimodal tasks. For instance, Llava~\cite{llava,llava1_5} introduces a linear projection layer between the vision encoder and the LLM to better align visual-language features. Subsequent works~\cite{qwenvl,mobilevlmv2,cogvlm} have further improved the performance of LVLMs through various architectures. However, the performance of these models in complex scenarios, particularly those involving camouflage objects, remains to be thoroughly explored. Besides, recent studies~\cite{visionllmv2,codlvlm,eval-cod-lvlm,mmt-bench} have evaluated the performance of LVLMs on the camouflaged object detection task, indicating that current LVLMs exhibit limitations in understanding and recognizing such complex scenes.

\section{Datasets and Benchmark}

\begin{table}[!t]
    \centering
    \small
    \begin{tabular}{lcc}
    \hline
    \toprule
         Source &Num. MM &Num. B\\ \midrule
         COD-10K~\cite{COD10K}&  4,765 &257\\ 
         MoCA-Mask~\cite{MoCA-Mask} &4,449 &242\\ 
         PlantCamo~\cite{PlantCamo} &1,181 &69\\ 
         CPD-1K~\cite{CPD1K} & 968 &32\\ \midrule 
         SUM &11,393 & 600 \\
    \bottomrule
    \hline
    \end{tabular}
    \caption{Image sources for the MM-CamObj dataset (Num. MM) and CamObj-Bench (Num. B).}
    \label{tab:datasource}
\end{table}

In our study, the images in the MM-CamObj dataset are sourced from publicly available camouflage scene understanding datasets. These datasets not only provide accurate category annotations but also include segmentation masks for camouflaged objects. Specifically, as shown in Table~\ref{tab:datasource}, we carefully select 11,963 camouflaged target images from~\cite{ovcamo,COD10K,MoCA-Mask,PlantCamo,CPD1K}. Of these, 11,363 images are used to construct CamObj-Align and CamObj-Instruct, while the remaining 600 images are utilized to construct CamObj-Bench.

\subsection{Metadata Construction}

We define a standardized metadata format for MM-CamObj and CamObj-Bench. Each metadata entry contains the following key information: the name of the source dataset, a unique identifier, the basic category of the camouflaged object, the RGB image, and its corresponding camouflaged object mask. Additionally, the metadata records the number of camouflaged objects, their bounding box information, and their proportion within the entire image. With this information, our constructed training samples can better align with the camouflaged scenes in the given images, allowing our benchmark to incorporate more diverse question types. Through a standardized metadata structure, researchers can more easily compare the capabilities of different models using a unified evaluation standard, thereby promoting the development of LVLMs in the domain of camouflaged scenes.

\subsection{CamObj-Align}

\begin{figure}
    \centering
    \includegraphics[width=0.9\linewidth]{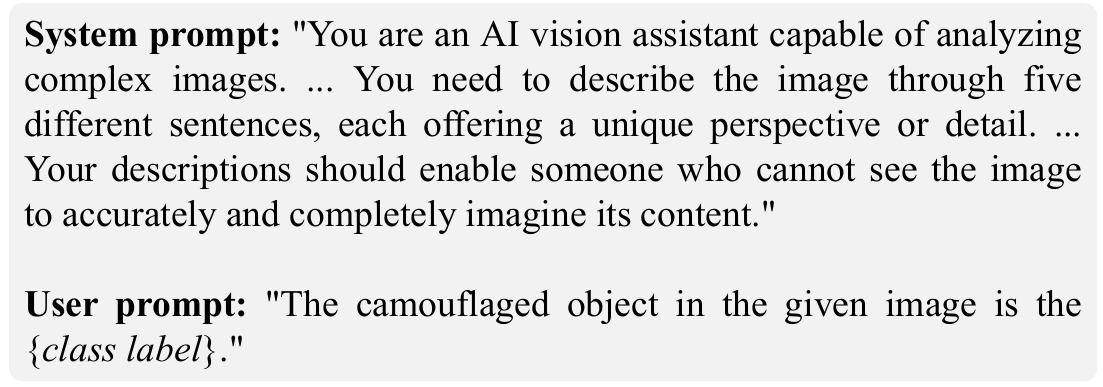}
    \caption{An example prompt snippet for constructing CamObj-Align.}
    \label{fig:pt_prompt}
\end{figure}

\begin{figure}
    \centering
    \includegraphics[width=0.9\linewidth]{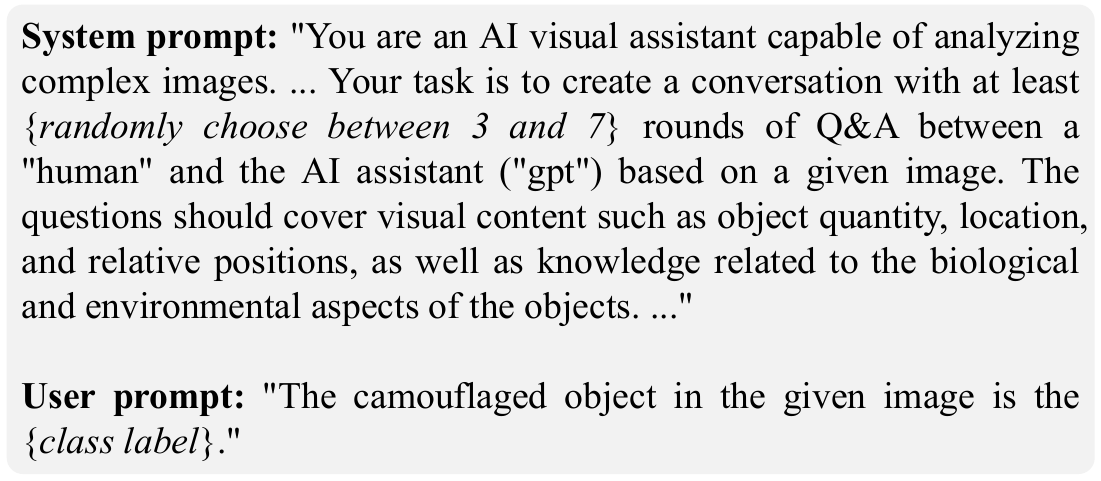}
    \caption{An example prompt snippet for constructing CamObj-Instruct.}
    \label{fig:sft_prompt}
\end{figure}

\begin{figure*}
    \centering
    \includegraphics[width=0.98\linewidth]{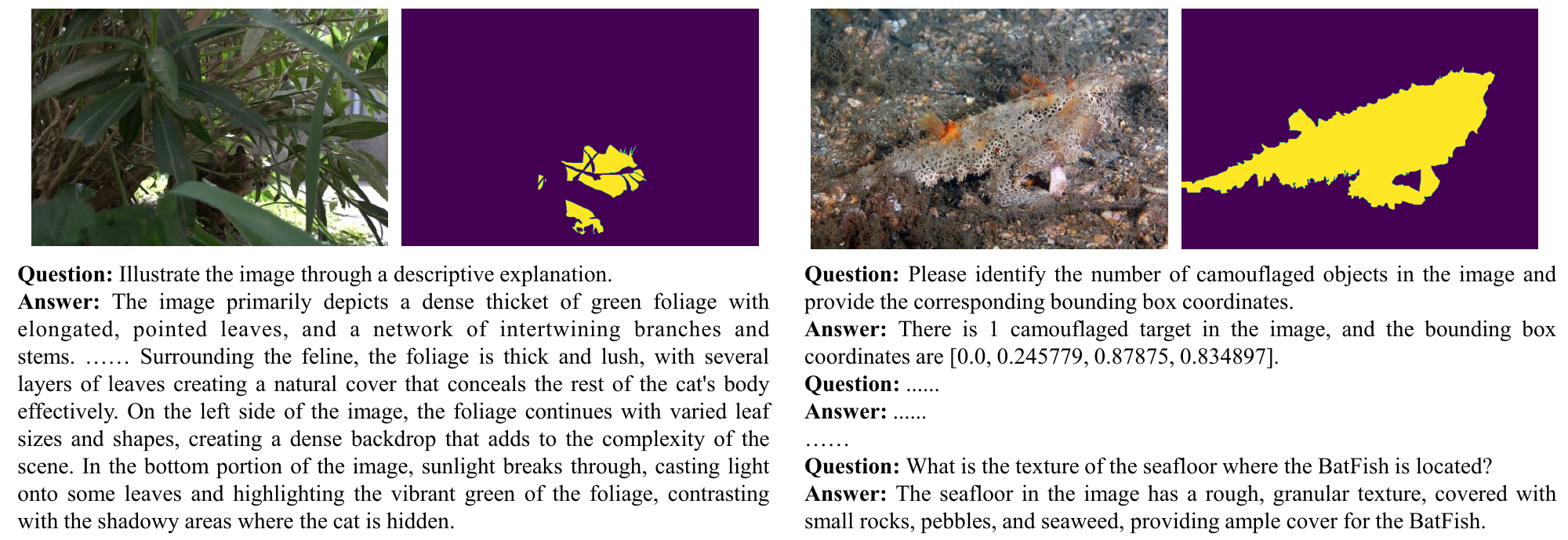}
    \caption{\textit{Left: }A sample example snippet in CamObj-Align. \textit{Right: }A sample example snippet in CamObj-Instruct. The corresponding mask images are solely for better visualization of the camouflaged objects. These masks are not included in the CamObj-Align and CamObj-Instruct datasets and will not be used during the training process.}
    \label{fig:example}
\end{figure*}

CamObj-Align dataset consists of 11,363 camouflaged object images along with their corresponding descriptions. To ensure the accuracy of these descriptions, we first use the advanced GPT-4o to generate detailed descriptions of camouflaged objects from five different perspectives, as shown in Figure~\ref{fig:pt_prompt}. We then conduct manual reviews and corrections to address errors in the descriptions, such as mistakes in object location and count. Finally, we adopt the instructions from~\cite{llava} as the detailed image description instructions. An example from CamObj-Align is presented in Figure~\ref{fig:example} Left. For more information on the prompts used in GPT-4o, as well as statistical information and sample examples of CamObj-Align, please refer to the Appendix.

\subsection{CamObj-Instruct}

CamObj-Align is utilized during the alignment phase of LVLMs to incorporate knowledge about camouflage objects and scenes into the model. To further enhance the model's ability to follow instructions, we construct the CamObj-Instruct dataset using 11,363 images and their corresponding camouflage object masks. To ensure diversity and complexity in the instructions, we use GPT-4o to generate dialogues, generating instructions and corresponding responses, as shown in Figure~\ref{fig:sft_prompt}. Subsequently, to ensure the accuracy of the dataset, we also conduct manual reviews to correct any hallucinated instructions and responses. Finally, leveraging the information of metadata, we introduce counting and localization instructions and responses in the dialogues to enhance LVLM's ability to perceive camouflage objects. As illustrated in Figure~\ref{fig:example} Right, we present an example from CamObj-Instruct dataset. More details on the GPT-4o prompts, statistics of CamObj-Instruct, and sample examples can be found in the Appendix.


\subsection{CamObj-Bench}
\subsubsection{Statistical Information}

To comprehensively evaluate the capabilities of LVLMs in camouflage scene tasks, we propose CamObj-Bench. This benchmark comprises 600 meticulously selected images along with their corresponding camouflage object masks, covering 144 different biological species. As shown in Table \ref{tab:datasource}, these images are sourced from four major public datasets: COD-10K \cite{COD10K}, MoCA-Mask \cite{MoCA-Mask}, PlantCamo \cite{PlantCamo}, and CPD-1K \cite{CPD1K}.


CamObj-Bench exhibits significant diversity in the number of camouflaged objects, presenting varying levels of challenges to LVLMs. Specifically, 83.7\% of the images contain one camouflaged object, 8.8\% contain two objects, 2.3\% contain three objects, 1.8\% contain four objects, and 3.3\% contain more than four camouflaged objects. Additionally, there is notable variation in the size of camouflaged objects within the images. Among these, 69.3\% of the objects occupy 0\% to 10\% of the image area, 18.0\% occupy 10\% to 20\%, 7.5\% occupy 20\% to 30\%, and 5.2\% occupy more than 30\% of the image area. These variations in relative size further complicate the recognition task. More detailed information about the benchmark can be found in the Appendix.

\subsubsection{Tasks Construction}

\begin{figure*}
    \centering
    \includegraphics[width=0.99\linewidth]{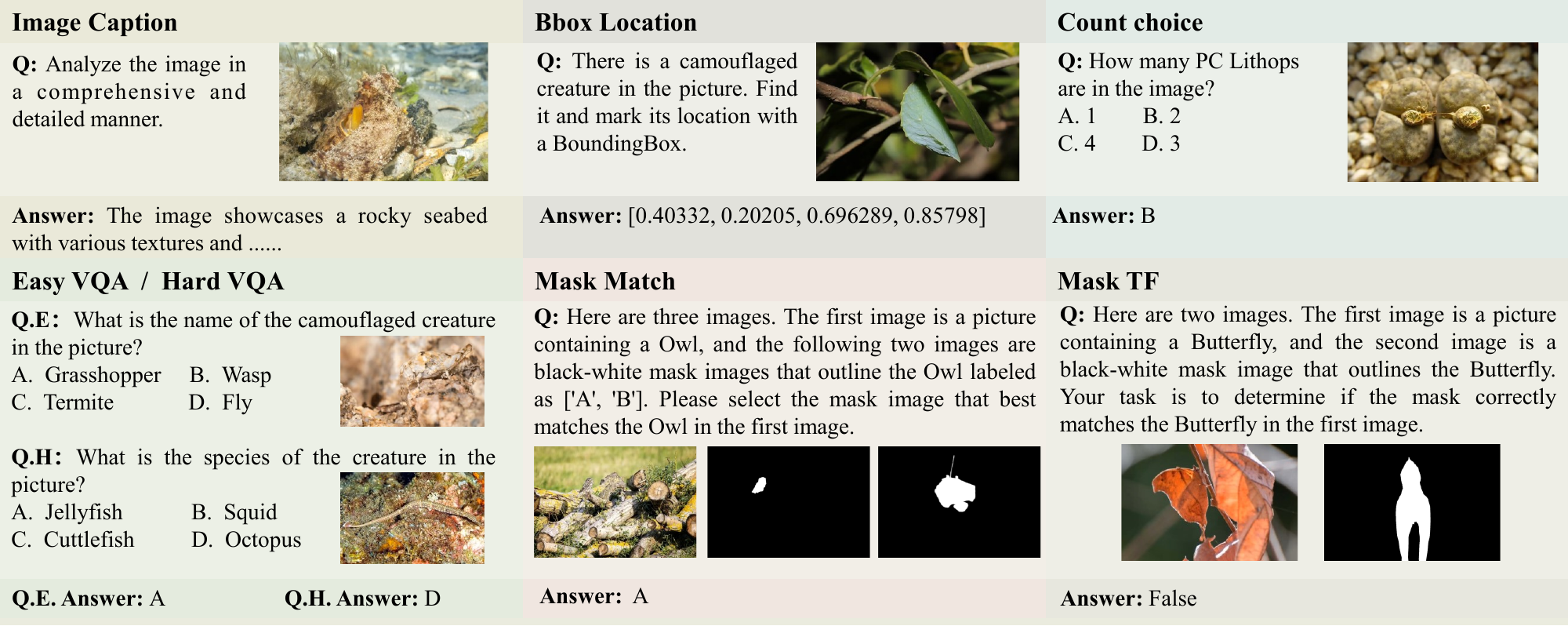}
    \caption{Some typical examples from the CamObj-Bench. Our benchmark consists of seven tasks to evaluate LVLMs' capabilities in recognition, classification, localization, counting, and scene understanding in the camouflaged object scenarios.}
    \label{fig:questions}
\end{figure*}

We have designed seven tasks in CamObj-Bench based on metadata information. As shown in Figure~\ref{fig:questions}, these tasks include Easy-VQA, Hard-VQA, Bbox Location, Image Caption, Count Choice, Mask Match, and Mask TF. The number of images and questions for each task are shown in Table~\ref{tab:task_counts}. The first five tasks utilize single-image input, while the last two involve multi-image input.

\textbf{Easy VQA}: Given a camouflaged scene image, the LVLM is required to select the name of the object from four options. The distractor options are generated by GPT-4o and are distinctly different from the correct answer. For example, if the correct answer is `Spider', the distractor options might be `Bird', `Frog', and `Fish'. Accuracy is used as the evaluation metric for this task.

\textbf{Hard VQA}: Similar to Easy VQA, but with more challenging distractor options that are closer to the correct answer. For instance, if the correct answer is `Spider', the distractor options might be `Water Spider', `Dragonfly', and `Mosquito', making it more difficult for LVLMs to correctly identify the object. This task also uses accuracy as the evaluation metric.

\textbf{Bbox Location}: Given an image of a camouflaged object, the LVLM needs to detect the location of the object and output it in the format of a bounding box \([x1, y1, x2, y2]\), where \(x1, y1\) represent the coordinates of the top-left corner, and \(x2, y2\) represent the coordinates of the bottom-right corner. Mean Intersection over Union (mIoU) is used as the evaluation metric for this task.

\begin{table}[!t]
\centering\footnotesize
\begin{tabular}{lcc}
    \hline
\toprule
\textbf{Task} & \textbf{\#Images} & \textbf{\#Questions} \\ \midrule
Easy-VQA      & 600                        & 3,000                         \\
Hard-VQA      & 599                        & 2,995                         \\
Bbox Location & 482                        & 482                          \\
Image Caption & 600                        & 600                          \\
Count Choice  & 572                        & 572                          \\
Mask Match    & 600                        & 600                          \\
Mask TF       & 600                        & 1,200                         \\ \bottomrule
    \hline
\end{tabular}
\caption{Image and Question Counts for CamObj-Bench.}
\label{tab:task_counts}
\end{table}

\textbf{Image Caption}: In this task, the LVLM is required to provide a detailed description of the image based on the camouflaged scene. The ground truth answers are generated by GPT-4o and Gemini-1.5-pro \cite{gemini15}. Subsequently, we manually inspect and combine the two responses, correcting the hallucinations. We employ BGE-v1.5-en \cite{BGE} to calculate the CLIPScore between the image and the description, using this as the evaluation metric.

\textbf{Count Choice}: This task also employs a single-choice format, where the LVLM is given an image and must identify the number of camouflaged objects within the image and select the correct answer from four options. Accuracy is used as the evaluation metric for this task.

\textbf{Mask Match}: Given an image of a camouflaged object and two mask images, where one mask is correct and the other is randomly selected and incorrect, the LVLM is required to select the correct mask image.

\textbf{Mask TF}: Given a camouflaged object image and a mask, the LVLM is required to determine whether the contour in the mask image matches the camouflaged object in the original image, answering either True or False. For each image in CamObj-Bench, we generate one `True' question with the correct mask and one `False' question with an incorrect, randomly selected mask.

\section{Methods}

Due to the challenging of acquiring high-quality camouflage images, our CamObj-Align and CamObj-Instruct, contain a limited number of training samples compared to general domain datasets. Therefore, in this section, we introduce a curriculum learning strategy to fully leverage the limited samples, enhancing the learning efficacy and improving the downstream performance of CamObj-Llava.

\subsection{Determination of Curriculum Learning Metrics}

The core idea of curriculum learning is to train models first on simpler samples and gradually increase the complexity samples, thereby improving the model's learning efficiency and generalization capability \cite{curriculum_learning_survey_1,curriculum_learning_survey_2}. This strategy simulates the human learning process, which progresses from simple to complex, helping to accelerate model convergence and enhance performance on complex tasks. However, determining the difficulty level of samples is a prerequisite for implementing curriculum learning.

In our CamObj-Align and CamObj-Instruct datasets, each sample consists of an image with the question-answer conversations between a human and an AI assistant. The `Question', posed by a human, can be regarded as an instruction, while the `Answer', provided by the AI assistant, can be considered a response. Consequently, we naturally employ the sentence lengths of instructions and responses as metrics to assess the difficulty level of the samples \cite{nlp_cl_1,nlp_cl_2}. Typically, longer sentence lengths are viewed as indicative of more challenging samples \cite{curriculum_learning_survey_1}. We examine three modes: instruction length ($\textup{Len}_{ins}$), response length ($\textup{Len}_{rsp}$), and their sum ($\textup{Len}_{all}$). Additionally, we introduce image-text similarity as an extra difficulty metric, where higher image-text similarity often implies lower sample difficulty. We consider the similarity between the instruction and image ($\textup{Sim}_{ins}$), the response and image ($\textup{Sim}_{rsp}$), and the entire dialogue and image ($\textup{Sim}_{all}$). We utilize BGE-M3 \cite{bge_m3} and BGE-v1.5-en \cite{BGE} to obtain embeddings for the image and text, calculating the cosine similarity between them. Subsequently, we average the similarity scores from these two sets of BGE models to measure the final image-text similarity.

\begin{table}[!t]
\centering\footnotesize
\begin{tabular}{lccc}
    \hline
\toprule
\textbf{Mode} & \textbf{Easy VQA} & \textbf{Hard VQA} & \textbf{Bbox Location} \\
\midrule

- &90.3 &68.9 &52.8  \\
\midrule
$\textup{Len}_{ins}$  &90.9&67.8&52.2 \\
$\textup{Len}_{rsp}$  &\underline{91.3}&\textbf{70.6}&52.8 \\
$\textup{Len}_{all}$  &90.9&69.8&\textbf{56.7} \\
\midrule
$\textup{Sim}_{ins}$  &90.8&\underline{70.3}&51.8 \\
$\textup{Sim}_{rsp}$  &\underline{91.3}&69.1&54.7 \\
$\textup{Sim}_{all}$  &\textbf{91.5}&69.3&\underline{54.9} \\
\bottomrule
    \hline
\end{tabular}
\caption{Preliminary studies with various curriculum learning modes. Bold indicates the best, underline indicates the second best.}
\label{tab:preexp}
\end{table}

\begin{table*}[!t]
\centering\footnotesize
\begin{tabular}{l|c|c|c|c|c|c}
    \hline
\toprule
\textbf{Model} &\textbf{Size} & \textbf{Easy VQA} & \textbf{Hard VQA} & \textbf{Bbox Location} & \textbf{Image Caption} & \textbf{Count choice} \\
\midrule
\hline
\multicolumn{7}{l}{\emph{Open-source LVLMs}}\\
\hline
Fuyu &8B& 35.9 & 26.1  & - & 69.1  & 24.1 \\
Idenfics &9B& 79.2 & 69.1  & - & 63.5  & 44.1 \\
InstructBLIP &11B& 72.0 & 55.0  & - & 68.2   & 68.4 \\
Kosmos2 &1.6B& 25.3 & 25.2  & - & 70.4   & 28.8 \\
Llava-v1.5-7B &7B& 78.2 & 59.0  & 37.4 & 74.4  & 70.5 \\
Llava-v1.5-13B &13B& 83.4 & 63.0  & 40.8 & 74.7   & 82.9 \\
MiniCPM-V-2.5 &8B& 66.1 & 41.2  & 33.3 & 77.1  & 78.7 \\
Qwen-VL-Chat &7B& 77.8 & 59.7  & 21.2 & 74.9   & 67.3 \\
CamObj-Llava-7B &7B& 90.9 & 69.8  & 56.7 & 85.0  & 81.5 \\
CamObj-Llava-13B &13B& 91.8 &71.9  & \textbf{59.4} & \textbf{85.7}  & \textbf{83.0}  \\
\midrule
\hline
\multicolumn{7}{l}{\emph{Closed-source LVLMs}}\\
\hline
GPT-4o &-& 88.5 & 79.9  & 32.0  & 81.8   & 70.8\\
GPT-4o-mini &-& 86.8 & 72.4  & 28.1 & 80.6   & 58.4 \\
Gemini-1.5-pro &-& \textbf{95.6} & \textbf{86.4}  & 45.0 & 79.5   & 65.3 \\
\bottomrule
    \hline
\end{tabular}
\caption{Validation results on single-image tasks in CamObj-Bench. We use accuracy (\%) as the evaluation metric for Easy VQA, Hard VQA, and Count choice, mIoU (\%) as the evaluation metric for Bbox Location, and CLIPScore (\%, extend the range from 0 to 1 into percentage units.) as the evaluation metric for Image Caption.}
\label{tab:single_image_tasks}
\end{table*}

\subsection{Preliminary Experiment}

After determining the difficulty level of the samples, we sort them from easy to difficult and gradually introduce them into the training process according to the principles of curriculum learning. Following the setup of Llava \cite{llava1_5}, we conduct a two-stage training on the Llava-v1.5-7b. First, during the alignment stage, we introduce knowledge related to camouflage scenes using the CamObj-Align dataset. Second, in the instruction fine-tuning stage, we enhance the model's ability to follow instructions with the CamObj-Instruct dataset. Different from Llava, we employ LoRA \cite{nlplora} fine-tuning method in both stages.

To validate the effectiveness of curriculum learning, we evaluate its performance on three tasks within the CamObj-Bench: Easy VQA, Hard VQA, and Bbox Location. As shown in Table \ref{tab:preexp}, introducing curriculum learning generally enhances model performance. For instance, in the $\textup{Len}_{all}$ mode, performance on the Bbox Location task improves by 3.9\% compared to the baseline without curriculum learning. However, in the $\textup{Len}_{ins}$ mode, performance on the Hard VQA and Bbox Location tasks is even lower than without curriculum learning. This outcome aligns with expectations because, similar to \cite{llava,llava1_5}, the human instructions used in the CamObj-Align dataset are randomly selected from a fixed template library, resulting in similar semantic information and sentence length between instructions. After comprehensive consideration, we choose $\textup{Len}_{all}$ as the training strategy for CamObj-Llava to maximize the utilization of limited training data and achieve better performance in downstream tasks.

\section{Experiments}

\subsection{Baseline}

In this section, we conduct a comprehensive evaluation of LVLMs' performance across the seven tasks of the CamObj-Bench. We test a range of open-source LVLMs, including Fuyu-8B \cite{fuyu8b}, Idenfics \cite{idenfics}, InstructBLIP-FlanT5-XXL \cite{InstructBLIP}, Kosmos2 \cite{kosmos2}, Llava-v1.5-7B \cite{llava1_5}, Llava-v1.5-13B \cite{llava1_5}, MiniCPM-Llama3-V-2.5 \cite{minicpm}, and Qwen-VL-Chat \cite{qwenvl}. In addition, we evaluate several advanced closed-source models such as GPT-4o\footnote{https://platform.openai.com/docs/models/gpt-4o}, GPT-4o-mini\footnote{https://platform.openai.com/docs/models/gpt-4o-mini}, and Gemini-1.5-pro \cite{gemini15}. The selection of these models aims to provide a comprehensive performance benchmark to analyze the performance of different models across various tasks.

\subsection{Training Details}

In our study, we build the CamObj-Llava-7B/13B by continuously training on the CamObj-Align and CamObj-Instruct datasets, based on the Llava-v1.5-7B/13B \cite{llava1_5}. The training consists of two stages: alignment and instruction fine-tuning. During the alignment stage, we set the learning rate to 5e-4, and for the instruction fine-tuning stage, the learning rate is adjusted to 2.5e-4. Both stages are trained for 1 epoch using the AdamW \cite{adamw} optimizer, with a cosine decay strategy \cite{cosineannealingLR} to dynamically adjust the learning rate. Specifically, we apply LoRA \cite{nlplora} modules to each linear layer of the large language model, with the rank \( r \) and scaling factor \( \alpha \) set to 128 and 256. All experiments on conducted on 8×NVIDIA A800 GPUs.

\begin{table}[!t]
\centering
\footnotesize
\begin{tabular}{l|c|c|c}
    \hline
\toprule
\textbf{Model} & \textbf{Mode} & \textbf{Mask Match} & \textbf{Mask TF} \\
\midrule
\hline
\multicolumn{4}{l}{\emph{Open-source LVLMs}}\\
\hline
Fuyu &merge & 51.8 & 50.4\\
Idenfics &sequence & 43.1 & 47.4 \\
InstructBLIP &merge & 43.2 & 47.4 \\
Kosmos2 &merge & 50.0 & 50.0 \\
Llava-v1.5-7B &merge & 50.8 & 50.0 \\
Llava-v1.5-13B &merge & 45.7 & 47.0 \\
MiniCPM-V-2.5 &sequence & 51.7 & 48.3 \\
Qwen-VL-Chat &merge & 50.0 & 50.0  \\
CamObj-Llava-7B &merge & 50.0 & 50.0 \\
CamObj-Llava-13B &merge & 50.0 &49.5 \\
\midrule
\hline
\multicolumn{4}{l}{\emph{Closed-source LVLMs}}\\
\hline
GPT-4o &sequence & \textbf{95.8} & \textbf{88.3} \\
GPT-4o-mini &sequence  & 84.8 & 69.0 \\
Gemini-1.5-pro &sequence & 87.8 & 72.0 \\
\bottomrule
    \hline
\end{tabular}
\caption{Validation results on multiple-images tasks in CamObj-Bench. We use accuracy (\%) as the evaluation metric for Mask Match and Mask TF.}
\label{tab:multi_image_tasks}
\end{table}

\subsection{Main Results}


\subsubsection{Single-image Tasks}

Table \ref{tab:single_image_tasks} shows the performance of existing LVLMs and our CamObj-Llava on five single-image tasks. Our CamObj-Llava-7B/13B models, trained with CamObj-Align and CamObj-Instruct, outperform all other open-source models in these tasks. Compared to Llava-v1.5-7B/13B, our CamObj-Llava-7B/13B achieves an average improvement of 23.2\% and 16.9\% across these tasks.

\textbf{CamObj-Llava excels in camouflaged object recognition.} In the Easy and Hard VQA tasks, we evaluate the ability of LVLMs to recognize and classify camouflaged objects. Although our model underperforms compared to the closed-source models in the Hard VQA task, CamObj-Llava-7B surpasses GPT-4o by 2.4\% in the Easy VQA task, with CamObj-Llava-13B further improving upon this by 0.9\%.

\textbf{CamObj-Llava surpasses closed-source LVLMs in object localization, scene understanding, and counting.} In the Bbox Location task, which primarily evaluates the localization ability of LVLMs for camouflaged objects, our CamObj-Llava-7B/13B outperforms all advanced closed-source models. Compared to the top-performing Gemini-1.5-pro, CamObj-Llava-13B achieves a 14.4\% improvement in the mIoU metric. The Image Caption task focuses on assessing the model's understanding of camouflaged scenes. In this task, CamObj-Llava-7B/13B also outperforms leading closed-source models. Specifically, CamObj-Llava-7B, with only 7B parameters, surpasses GPT-4o by 3.2\% in CLIPScore. The Count choice task evaluates the counting abilities of LVLMs, and once again, our CamObj-Llava exceeds all closed-source models.

The results across these five single-image tasks demonstrate that our CamObj-Align and CamObj-Instruct datasets provide comprehensive knowledge of camouflaged objects and scenes, significantly enhancing LVLMs' capabilities in recognition, classification, localization, counting, and scene understanding.

\subsubsection{Multiple-images tasks}

In handling multiple-images tasks, we draw inspiration from previous research \cite{mantis} and employ two different processing methods, depending on whether the LVLM supports multiple-images input. The first strategy, termed `merge', involves horizontally concatenating multiple images into a single image for processing. The second strategy, termed `sequence', involves inputting multiple images in a sequential manner.

Table \ref{tab:multi_image_tasks} shows the performance of existing LVLMs and our CamObj-Llava on the Mask Match and Mask TF tasks. For open-source LVLMs, including our model, the performance on these multiple-images tasks is close to random selection. This phenomenon could also be observed in \cite{blink,mantis}. One possible reason is that during training, only a few or no multiple-images input modes and related instructions are used, resulting in poor performance on multiple-images tasks. In contrast, advanced closed-source models have better adaptability to multiple-images tasks. However, even the advanced GPT-4o-mini achieves only 69.0\% accuracy on the Mask TF task, only 19.0\% higher than random selection. This further indicates that our CamObj-Bench presents new challenges for existing LVLMs. The case studies are in the Appendix.

\subsection{Ablation Studies}

To validate the performance of our CamObj-Llava-7B with limited data, we randomly selected 10\%, 20\%, and 50\% of the samples from the CamObj-Align and CamObj-Instruct datasets for training, while keeping the rest of the experimental settings consistent with those described in Sec. `Training Details'. The results are shown in Table \ref{tab:ablation}. Even with only 10\% of the data samples, CamObj-Llava-7B surpasses Llava-v1.5-7B in the Easy VQA, Hard VQA, and Bbox Location tasks, with improvements of 6.4\%, 2.7\%, and 5.3\%, respectively. This demonstrates the significant advantage of our training dataset in providing LVLMs with rich knowledge about camouflage scenarios.

\begin{table}[!t]
\centering
\footnotesize
\begin{tabular}{lccc}
    \hline
\toprule
\textbf{Ratio} & \textbf{Easy VQA} & \textbf{Hard VQA} & \textbf{Bbox Location} \\
\midrule
- &78.2&59.0&37.4 \\
\midrule
10\%  &84.6&61.7&42.7 \\
20\%  &86.0&63.9&44.2 \\
50\%  &87.8&65.5&49.4 \\
100\%  &\textbf{90.9}&\textbf{69.8}&\textbf{56.7} \\
\bottomrule
    \hline
\end{tabular}
\caption{Ablation studies on the Easy/Hard VQA and Bbox Location tasks with various training samples. Ratio refers to the proportion of training samples randomly selected from CamObj-Align and CamObj-Instruct datasets. `-' stands for Llava-v1.5-7B without training on our datasets.}
\label{tab:ablation}
\end{table}
\section{Conclusion}

In this paper, we aim to enhance the capabilities of large vision-language models in complex scenarios, particularly those involving camouflaged objects. To achieve this, we have collected and constructed the MM-CamObj dataset, which includes CamObj-Align for alignment training and CamObj-Instruct for instruction fine-tuning. Additionally, to effectively utilize the limited training samples, we introduce six different modes of curriculum learning and continue training based on Llava-v1.5, resulting in our CamObj-Llava. To comprehensively validate the performance of existing LVLMs and our model in camouflaged scenarios, we develop CamObj-Bench, a benchmark consisting of seven challenging tasks. Extensive experiments demonstrate that our CamObj-Llava surpasses the advanced GPT-4o in four of these tasks. In the future, we plan to expand the dataset and introduce multi-image training for our CamObj-Llava.

\bibliography{aaai25}
\end{document}